%% file: main.tex
\documentclass[10pt,twocolumn,letterpaper]{article}

\usepackage{cvpr}

\input{preamble}

\definecolor{cvprblue}{rgb}{0.21,0.49,0.74}
\usepackage[pagebackref,breaklinks,colorlinks,citecolor=cvprblue]{hyperref}

\usepackage{amssymb}
\usepackage{float}
\usepackage{graphicx}
\usepackage[table]{xcolor}

\usepackage{multirow} 
\usepackage[accsupp]{axessibility}

\def\paperID{****} 
\def\confName{CVPR}
\def\confYear{2026}

\title{FusionRegister: Every Infrared and Visible Image Fusion Deserves Registration}

\author{
    Congcong Bian$^{1}$ \quad 
    Haolong Ma$^{1}$ \quad 
    Hui Li$^{1}$\thanks{Corresponding author} \quad 
    Zhongwei Shen$^{2}$ \\ 
    Xiaoqing Luo$^{1}$ \quad 
    Xiaoning Song$^{1}$ \quad 
    Xiao-jun Wu$^{1}$ \\
    \noalign{\vspace{3pt}}
    \small $^{1}$ School of Artificial Intelligence and Computer Science, Jiangnan University, Wuxi, China \\
    \small $^{2}$ School of Electronic and Information Engineering, Suzhou University of Science and Technology, Suzhou, China \\
    \tt\small bociic\_jnu\_cv@163.com, ninesxd@qq.com, \{lihui.cv, xqluo, x.song, wu\_xiaojun\}@jiangnan.edu.cn \\
    {\tt\small shenzw@usts.edu.cn}
}

\begin{document}
\maketitle
\input{abstract}    
\input{intro}

\input{relate}

\input{method}

\input{exp}

\input{conclusion}

\section{Acknowledgement}
This work is supported by the National Key Research and Development Program of China (2023YFE0116300), the National Natural Science Foundation of China (62202205, U25A20527, 62306203), the National Key Research and Development Program of China
under Grant (2023YFF1105102, 2023YFF1105105)

{
    \small
    \bibliographystyle{ieeenat_fullname}
    \bibliography{main}
}



\end{document}

%% file: preamble.tex
\usepackage[dvipsnames]{xcolor}


%% file: abstract.tex
\begin{abstract}
Spatial registration across different visual modalities is a critical but formidable step in multi-modality image fusion for real-world perception. Although several methods are proposed to address this issue, the existing registration-based fusion methods typically require extensive pre-registration operations, limiting their efficiency. 
To overcome these limitations, a general cross-modality registration method guided by visual priors is proposed for infrared and visible image fusion task, termed FusionRegister.
Firstly, FusionRegister achieves robustness by learning cross-modality misregistration representations rather than forcing alignment of all differences, ensuring stable outputs even under challenging input conditions.
Moreover, FusionRegister demonstrates strong generality by operating directly on fused results, where misregistration is explicitly represented and effectively handled, enabling seamless integration with diverse fusion methods while preserving their intrinsic properties. 
In addition, its efficiency is further enhanced by serving the backbone fusion method as a natural visual prior provider, which guides the registration process to focus only on mismatch regions, thereby avoiding redundant operations. 
Extensive experiments on three datasets demonstrate that FusionRegister not only inherits the fusion quality of state-of-the-art methods, but also delivers superior detail alignment and robustness, making it highly suitable for infrared and visible image fusion method.
The code will be available at \href{https://github.com/bociic/FusionRegister}{https://github.com/bociic/FusionRegister}.

\end{abstract}

%% file: intro.tex
\section{Introduction}
\label{sec:intro}
In recent years, Infrared and Visible Image Fusion (IVIF) has made significant progress. Non-deep learning methods achieve efficient information selection and adaptive fusion \cite{MS-2,hybird,sparse-li,subspace}. Deep learning-based approaches further exploit the power of neural models to extract both shared and modality-unique features for more accurate fusion \cite{li2018densefuse,ma2019fusiongan,rfnnest,lrrnet}. Modality-shared information refers to features that are clearly present in both modalities, while modality-unique information represents those observable only in one modality \cite{occo}. However, due to limitations of imaging devices, infrared and visible images are often misaligned. Direct fusion of unregistered images leads to severe information displacement, degrading fusion quality and overall efficiency.

\begin{figure}[!tbp]
\centering
\includegraphics[width=1\linewidth]{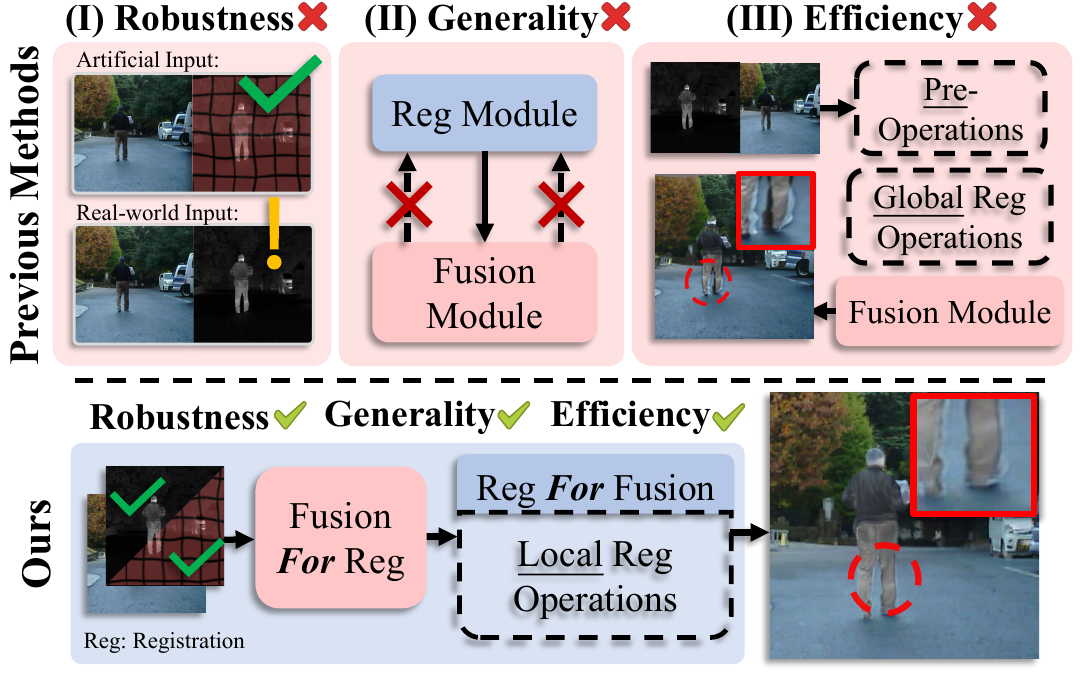}
\caption{Existing registration-based fusion methods face three key limitations: (\uppercase\expandafter{\romannumeral1})  Depend on artificial deformations; (\uppercase\expandafter{\romannumeral2}) Cannot interact with fusion methods; (\uppercase\expandafter{\romannumeral3}) Require extensive pre-operations and global registration. We propose a visual prior-based post-registration method that leverages fusion results to preserve fusion quality while significantly enhancing structural accuracy, achieving efficiency and robustness.}
\vspace{-0.4cm}
\label{fig:1}
\end{figure}

To address the above drawback, many studies have focused on registration as a pre-processing step. Existing methods, including image- or feature-level registration, utilize style transfer to reduce modality discrepancies, simplifying optical flow prediction \cite{umfusion,Wang_2024_IMF}. However, style transfer often incurs high computational cost and inevitable information loss. Other approaches map modality-specific features into a latent shared space to extract aligned representations \cite{murf,xu2022rfnet}. More recent works predict offset parameters directly in the intermediate feature domain, enabling one-stage fusion of aligned features \cite{ivfwsr,rfvif}. Nevertheless, as shown in Fig.~\ref{fig:1}, most existing cross-modal registration strategies still depend on computationally intensive pre-alignment operations to disentangle and align modal information, which limits their generality and computational efficiency.

\begin{figure}[!tbp]
\centering
\includegraphics[width=1\linewidth]{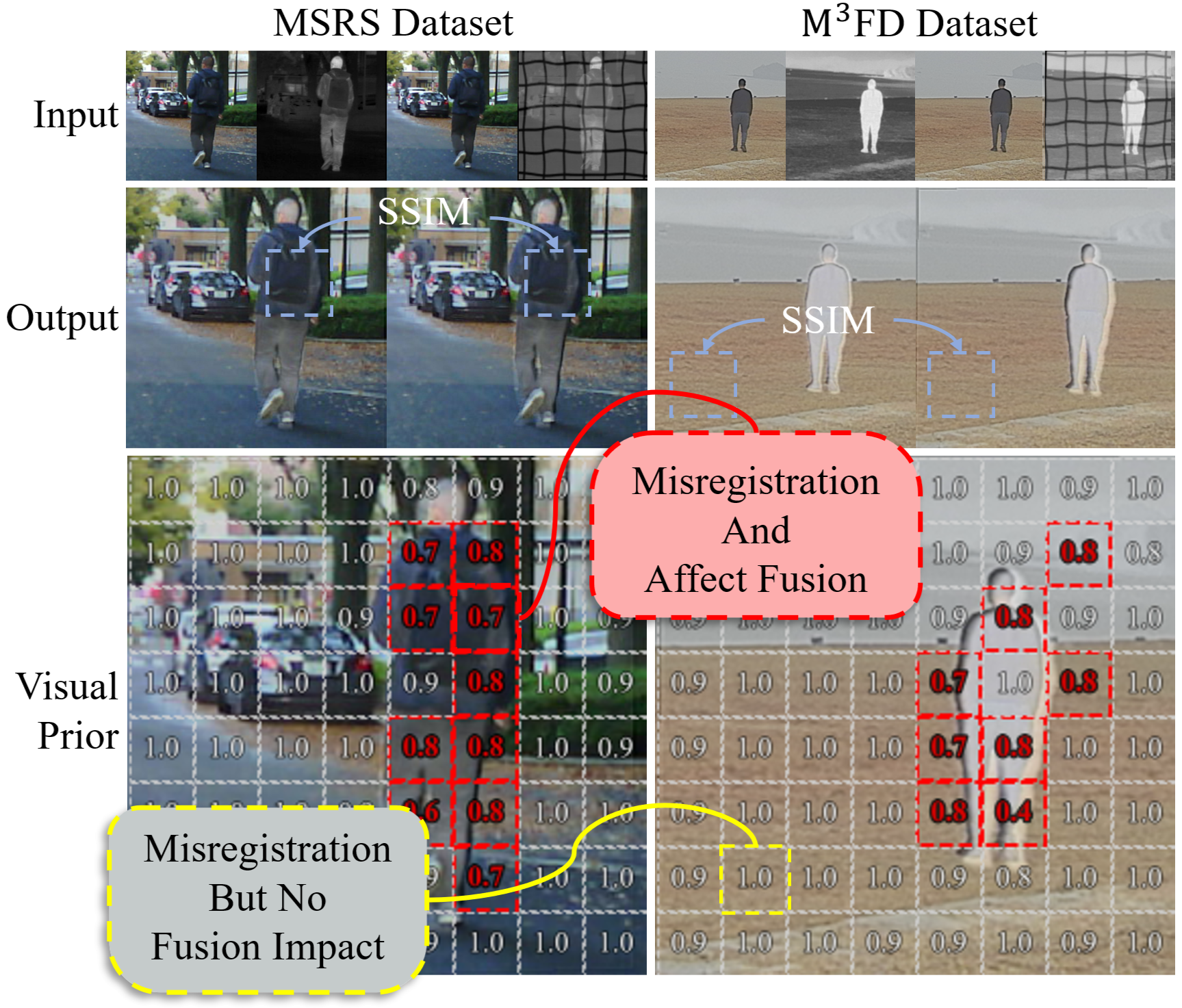}
\caption{The fusion results show that even with spatial deformation in the infrared image, misregistration appears only in modality-shared regions, where structural similarity with the fine registered result is significantly reduced.
}
\vspace{-0.6cm}
\label{fig:ssim}
\end{figure}

Moreover, most registration-based fusion methods attempt to align all information from both modalities during fusion, ignoring a key fact of image fusion: \textbf{NOT ALL features are preserved in the fused image}. As shown in Fig.~\ref{fig:ssim}, we take two perfectly registered pairs of a visible image $I_{vi}$ and an infrared image $I_{ir}$, and apply a spatial transform to create two misaligned pairs $I_{vi}$ and $I_{ir}^\phi$. All pairs are processed by the same image fusion backbone \cite{mmdrfuse} to generate fused results $I_f$ and $I_f^\phi$. Patch-level similarity analysis shows that spatial misregistration in the infrared image mainly affects the modality-shared regions, while the modality-specific regions remain mostly unchanged. This suggests that post-registration is only needed in the misaligned regions, ensuring accurate fusion without redundant global registration or heavy pre-processing. Based on this visual prior, we design FusionRegister, a universal post-registration framework as shown in Fig.~\ref{fig:1}, which balances robustness, generality, and efficiency.

Furthermore, multi-modal sensors are typically positioned in close proximity, naturally providing coarse registration between modalities. This implies that registration–joint fusion methods should be capable of directly handling raw sensor inputs. Nevertheless, many state-of-the-art methods rely on artificially generated deformations as supervisory signals, forcing networks to learn registration parameters from synthetic perturbations. As a result, these methods tend to collapse when confronted with real-world inputs that lack such deformations.

In contrast, our method focuses on learning and localizing misregistration representations, thereby enabling targeted correction within affected regions. This design enhances the robustness of fusion methods, allowing the fusion network to perform reliably on both deformed and clean inputs.

Moreover, evaluating registration quality in IVIF also remains challenging due to the lack of perfectly aligned reference pairs. Existing metrics rely on synthetic distortions, which fail to reflect real-world misregistration. To enable fair and structure-aware evaluation, we employ the Segment Anything Model (SAM)~\cite{sam} to obtain unbiased structural masks. This combination provides a consistent and fine-grained measure of registration accuracy.
The main contributions of this work are summarized as follows:
\begin{itemize}

\item A novel post-registration paradigm using visual priors that preserves fusion quality while significantly improving registration accuracy is proposed, demonstrating strong generality across diverse fusion methods.

\item A general framework seamlessly integrated with various infrared-visible fusion approaches is introduced, efficiently addressing cross-modal misregistration with minimal computational overhead.

\item A robust misrepresentation learning mechanism tailored to real-world scenarios is designed, capturing diverse misregistration representation to enhance adaptability across challenging conditions.

\end{itemize}


%% file: relate.tex
\begin{figure*}[!tbp]
\centering
\includegraphics[width=1\linewidth]{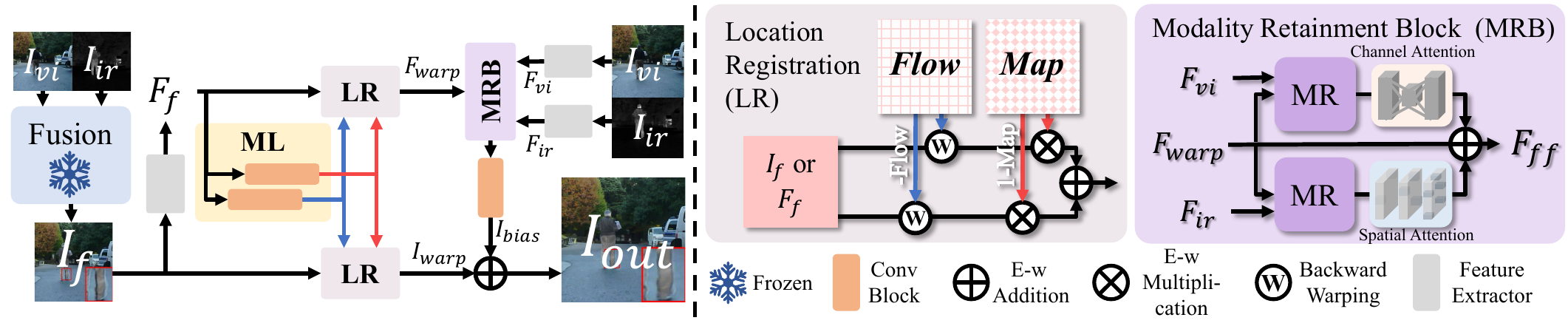}
\caption{The proposed FusionRegister consists of two core components: post-registration of misregistration regions via bi-directional warping, and a modality retainment block (MRB). Note that ML is misregistration location operation and e-w denotes element-wise.}
\label{fig:framework}
\end{figure*}
\section{Related Work}
\subsection{Infrared and Visible Image Fusion}

Early deep learning-based IVIF methods mainly rely on CNNs for feature extraction and fusion. Liu et al.~\cite{first_cnn} apply CNNs to this task, revealing the potential of end-to-end learning but suffering from limited representation and high complexity. To improve efficiency, \cite{lrrnet, mmdrfuse} introduces low-rank representation and distillation. AE-based frameworks \cite{li2018densefuse, li2020nestfuse, rfnnest, crossfuse} are then proposed to enhance multi-scale and structural representation and progressively improve feature interaction and reconstruction quality. 

To reduce dependence on paired data and improve perceptual realism, GAN-based methods introduce unsupervised adversarial learning. FusionGAN~\cite{ma2019fusiongan} pioneers this direction, and DDcGAN~\cite{ma2020ddcgan} enhances detail sharpness through dual discriminators, inspiring attention-~\cite{attentiongan}, frequency-~\cite{freqgan}, and gradient-based~\cite{tgfuse} variants. Transformer-based models \cite{ift, cdd, textfusion, promptfusion, gifnet, han2025cross} further advance IVIF by modeling long-range dependencies and global semantics and extend fusion via textual and prompt guidance to resolve semantic ambiguity.

Recently, diffusion and state-space models push IVIF toward higher fidelity and scalability. \cite{dif-fusion, ddfm, comofusion, Tang2025ControlFusion, Tang2024DRMF, han2025hi} improve consistency through diffusion generation. S4Fusion~\cite{s4fusion} and MambaDFuse~\cite{li2024mambadfuse} further achieve lightweight long-range modeling with state-space dynamics. 

Based on the above discussions, in this work, five representative methods from these paradigms are used to evaluate the generalization of FusionRegister.

\subsection{Registration-joint IVIF}

In real world scenarios, the misalignment between multi-modal sensors remains a major challenge in IVIF, as imperfect alignment causes ghosting and distortion. Style transfer mitigate modality gaps \cite{nemar, Wang_2024_IMF}. \cite{xu2022rfnet, murf, reconet} try coarse-to-fine registration combining rigid and non-rigid stages. 

To avoid information bias caused by pre-operations, feature-level strategies such as IVFWSR~\cite{ivfwsr} and RFVIF~\cite{rfvif} directly align latent representations to improve fusion under misalignment, while MulFS-CAP~\cite{cap} and SemLA~\cite{semla} leverage modality dictionaries and semantic priors for more robust matching. 

However, most existing registration-fusion frameworks still follow a ``register-then-fuse'' paradigm, incurring high preprocessing cost and weak adaptability. In contrast, our approach performs fusion first and then corrects misaligned regions guided by visual priors, ensuring both robustness and efficiency with new fusion models.

%% file: method.tex
\section{Methodology}
\label{sec:method}

\subsection{Overview}

The proposed method, FusionRegister, aims to correct spatial misregistration that persists after image fusion while preserving modality-specific details. 

Given a visible image ($I_{vi}$), an infrared image ($I_{ir}$), and their fused result ($I_{f}$), FusionRegister refines the fusion by localizing and correcting regions affected by misregistration, guided by visual priors embedded between $I_f$ and $I_{gt}$. $I_{gt}$ is the perfectly registered fused image. 

As illustrated in Fig.~\ref{fig:framework}, FusionRegister follows a visual prior guided pipeline composed of three collaborative stages:
(1) \textit{Misregistration Localization (ML)}, which detects spatial inconsistencies and estimates deformation fields;
(2) \textit{Location Registration (LR)}, which performs bi-directional warping to realign misregistered regions without damaging well-aligned ones; 
(3) \textit{Modality Retainment Block (MRB)}, which compensates texture and structure losses introduced during spatial transformation.

\subsection{Misregistration Localization and 
Location Registration}
\label{sec:ml&lr}
FusionRegister processes inputs in a hierarchical manner inspired by MIMO-UNet~\cite{2l}.  
At each scale $i\in {\{0,\cdots,N-1\}}$, downsampled versions $I_{in}^i$ ($in \in \{f, vi, ir\}$) are extracted, and three feature extractors that have the same structure but different parameters produce multi-scale features:

\begin{equation}\label{equ:1}
	\begin{split}
		F_{in}^0, \cdots, F_{in}^{N-1} &= FE_{in}(I_{in}^0, \cdots, I_{in}^{N-1}), \\
        in &\in \{f, vi, ir\},
	\end{split}
\end{equation}
where $I_{in}^0$ is source image, and $I_{in}^i \in{\mathbb{R}^{B\times3\times(H/2^i) \times (W/2^i)}}$. Detailed extractor architecture is provided in the supplementary material.

\textbf{Misregistration Localization (ML)} estimates a probability map $M^i \in \mathbb{R}^{B \times 1 \times (H/2^i) \times (W/2^i)}$ and a deformation field $\phi^i \in \mathbb{R}^{B \times 2 \times (H/2^i) \times (W/2^i)}$ that jointly represent the location and magnitude of misregistration. 
The field $\phi^i$ is progressively refined from coarse to fine levels:
\begin{equation}
\label{eq:phi}
\phi^i = \phi^i \otimes(1\oplus 2 \otimes Up(\phi^{i+1})),
\end{equation}
where $\oplus$ and $\otimes$ denote element-wise addition and multiplication, respectively. $Up(\cdot)$ indicates bilinear upsampling.
The factor of ``$2$" in Eq.~\ref{eq:phi} serves as a scaling coefficient applied to the upsampled deformation field $\phi^{i+1}$ to compensate for the resolution mismatch and maintain the physical consistency. 
This hierarchical propagation ensures spatial coherence and suppresses isolated deformations.

Compared with existing deformation field-based registration-based fusion frameworks, FR abandons the reliance on globally supervised guidance. Instead, it leverages visual prior cues to adaptively capture spatial misregistration representation and infer localized deformation transformations, enabling a more robust and generalizable field-free supervised paradigm.
\begin{figure}[!tbp]
\centering
\includegraphics[width=1\linewidth]{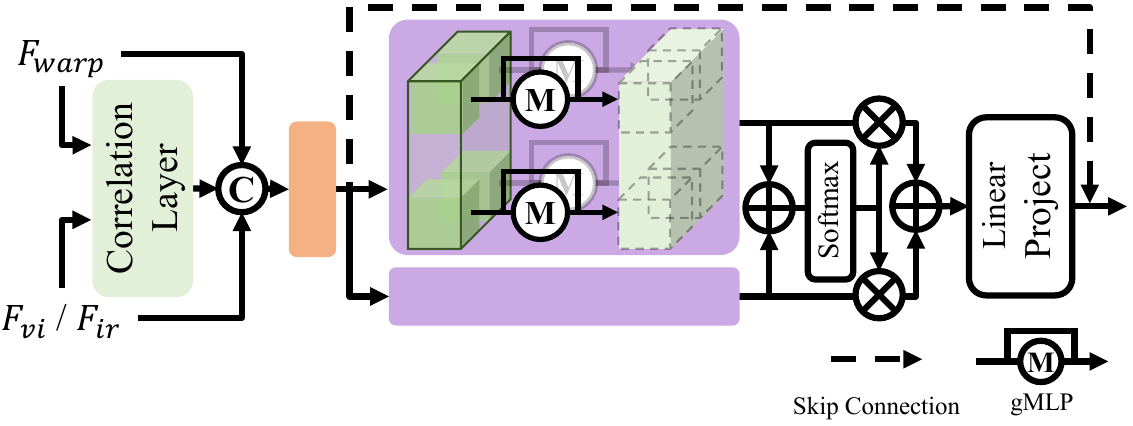}
\caption{Detailed architecture of the Modality Retainment Block (MRB).}
\label{fig:mrb}
\vspace{-0.2cm}
\end{figure}

\textbf{Location Registration (LR)} applies the predicted $\phi^i$ to correct the fused feature and fused image. 
Traditional single-direction backward warping may cause tearing or over-compensation. 
Thus, FusionRegister adopts a bi-directional warping strategy that symmetrically compensates distortions:
\begin{align}
\begin{split}
I_{\text{warp}}^i =& M^i \otimes BW(I_f^i,\phi^i)\quad\oplus \\
&  (1 - M^i) \otimes BW(I_f^i, -\phi^i),
\end{split} \\
\begin{split}
F_{\text{warp}}^i =& M^i \otimes BW(F_f^i,\phi^i)\quad  \oplus \\
& (1 - M^i) \otimes BW(F_f^i, -\phi^i),
\end{split}
\end{align}
where $BW(\cdot)$ denotes backward warping. 
The combination of forward and reverse corrections stabilizes deformation and prevents edge tearing. 
The resulting $F_{\text{warp}}^i$ is passed to MRB for detail restoration.

\subsection{Modality Retainment Block }
Spatial warping often weakens texture and attenuates contrast. To recover fine-grained information, we introduce the Modality Retainment Block (MRB). As shown in Fig.~\ref{fig:mrb}, a lightweight module leveraging gated MLP(gMLP)~\cite{gmlp} to enhance cross-modal feature retention.

\begin{figure}[!tbp]
\centering
\includegraphics[width=0.9\linewidth]{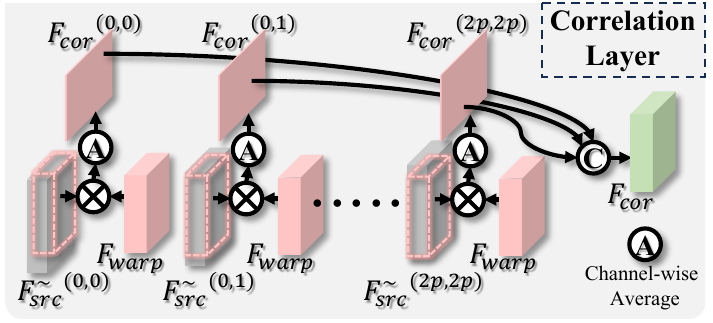}
\caption{Detailed architecture of the Correlation Layer.}
\vspace{-0.2cm}
\label{fig:cl}
\end{figure}

Firstly, a Correlation Layer~\cite{correlation} (Fig.~\ref{fig:cl}) measures local correspondence between the warped fusion feature $F_{warp}^i$ and source features $F_{src}^i$ ($src \in \{vi, ir\}$). 
By zero-padding $F_{src}^i$ with a range of $p$, resulting in a padded source feature ${\tilde{F}_{src}^{i}}\in \mathbb{R}^{B \times 2^iC \times (H/2^i+2p) \times (W/2^i+2p) }$. Then, ${\tilde{F}_{src}^{i,m,n}}\in \mathbb{R}^{B \times 2^iC \times (H/2^i) \times (W/2^i) }$ is the source feature map ${F_{src}^i}$ spatially shifted by $m$ pixels horizontally and $n$ pixels vertically in ${\tilde{F}_{src}^{i}}$, $m,n\in\{0,\dots,2p\}$. ${\tilde{F}_{src}^{i,m,n}}$ are compared with $F_{warp}^i$ to generate a correlation descriptor:

\begin{equation}
F_{cor}^{i,m,n}= CA({\tilde{F}_{src}^{i,m,n}}\otimes F_{warp}^i),
\end{equation}
where $CA(\cdot)$ denotes channel-wise averaging. All $F_{cor}^{i,m,n}$ are concatenated to form the final correlation feature $F_{cor}^i \in \mathbb{R}^{B \times 4p^2 \times (H/2^i) \times (W/2^i) }$. 
$F_{cor}^i$ serves as a bridge, recording local geometric relationships under multiple offsets. 
The concatenated features of $F_{warp}^i$, $F_{src}^i$, and $F_{cor}^i$ are compressed through a convolution block.

The compressed feature is partitioned into non-overlapping patches with different scales $s$. 
The gMLP models spatial interactions through channel mixing and gating, enabling efficient long-range dependency modeling without self-attention. For an input patch $F_{cor}^{i(s)} \in \mathbb{R}^{B\times 2^iC\times s\times s}$, gMLP applies a channel projection and a spatial gating unit:
\begin{equation}
F_{cor}^{i(s)}  = (F_{cor}^{i(s)}  W_1) \otimes \sigma((F_{cor}^{i(s)}  W_2) \mathbf{G}),
\end{equation}
where $W_1, W_2$ are linear projections, $\mathbf{G}$ is a learned gating matrix, 
$\sigma(\cdot)$ is ReLU function. 
All patches $F_{cor}^{i(s)}$ are concatenated back in their original spatial order to reconstruct the $F_{gMLP}^{i(s)}$. 
A softmax-weighted aggregation produces the final retained feature $F_{gMLP}^i$:
\begin{equation}
F_{gMLP}^i = \sum_{s} w_s \cdot F_{gMLP}^{i(s)}, \quad 
\end{equation}
where $w_s$ indicates the softmax weight of each scale $s$.

To emphasize modality-specific information, MRB integrates dual attention mechanisms:  

$\bullet$ \textit{Visible-modality attention} enhances semantic consistency through channel weighting:
\begin{equation}
\begin{split}
F_{f\&vi}^i =& F_{gMLP}^i \otimes \\
&(Sig\bigl(Conv(SA(F_{gMLP}^i))\bigr) \oplus 1),  
\end{split}
\end{equation}
where $SA(\cdot)$ denotes spatial average pooling and $Sig(\cdot)$ is the Sigmoid function.  

$\bullet$ \textit{Infrared-modality attention} emphasizes high-frequency details using spatial fusion:
\begin{equation}
\begin{split}
F_{f\&ir}^i =& F_{gMLP}^i \otimes \\
& (Sig\bigl(Conv(AM(F_{gMLP}^i))\bigr) \oplus 1),
\end{split}
\end{equation}
where $AM(\cdot)$ concatenates channel-wise max and mean responses.

The outputs are combined to form the refined feature:
\begin{equation}
 F_{ff}^i= F_{gMLP}^i\oplus F_{f\&vi}^i\oplus F_{f\&ir}^i.
\end{equation}

Finally, a convolution block predicts a residual bias map $I_{bias}^i \in \mathbb{R}^{B\times3\times(H/2^i)\times(W/2^i)}$, which refines the $I_{warp}^i$:
\begin{equation}
I_{out}^i = I_{warp}^i \oplus I_{bias}^i.
\end{equation}

\subsection{Loss Function}

The total loss jointly minimizes registration error while preserving structural and textural fidelity from both spatial and frequency domains, $\lambda_i$ balance the four components:
\begin{equation}
\mathcal{L}_{all} = \lambda_1 \mathcal{L}_{e} + \lambda_2 \mathcal{L}_{g} + \lambda_3 \mathcal{L}_{f} + \lambda_4 \mathcal{L}_{d}.
\end{equation}

Edge Loss ($\mathcal{L}_{e}$) aligns structural boundaries of the warped and fused results ($I_{warp}^i$, $I_{out}^i$) with the ground truth ($I_{gt}^i$), via a Difference of Gaussians (DoG) extractor $E(\cdot)$:


\begin{equation}
\begin{split}
&\mathcal{L}_{e} = \sum_{i=0}^{N-1}\sum_{j\in\{out, warp\}}\left\|(E(I^i_{j})\!-\!E(I_{gt}^i)\right\|_2, \\
&E(x) = x - G(Up(Down(G(x)))),
\end{split}
\end{equation}
here \( G(\cdot) \) denotes Gaussian filtering. The 1D Gaussian kernel is defined as: $k = [0.05,\ 0.25,\ 0.4,\ 0.25,\ 0.05]$. 
The 2D kernel is obtained by outer product \( G_k = k^\top k   \).

Global Spatial Loss ($\mathcal{L}_{g}$) constrains overall structural consistency in pixel space:
\begin{equation}
\mathcal{L}_{g} = \sum_{i=0}^{N-1}\!\left\| I_{out}^i - I_{gt}^i \right\|_2.
\end{equation}

Frequency Loss ($\mathcal{L}_{f}$) preserves high-frequency texture by minimizing Fourier-domain distance:
\begin{equation}
\mathcal{L}_{f} = \sum_{i=0}^{N-1}\!\left\| FFT(I_{out}^i) - FFT(I_{gt}^i) \right\|_1,
\end{equation}
where $FFT(\cdot)$ indicates the fast Fourier transform.

Detail Loss ($\mathcal{L}_{d}$) enhances texture consistency within misregistration map $M^i$, which is predicted in \ref{sec:ml&lr}, through the Sobel operator $\nabla(\cdot)$:
\begin{equation}
\mathcal{L}_{d} = \sum_{i=0}^{N-1}\!\left\| \nabla(I_{out}^i)\!\otimes\!M^i - \nabla(I_{gt}^i)\!\otimes\!M^i \right\|_1.
\end{equation}


\begin{figure}[!tbp]
\centering
\includegraphics[width=1\linewidth]{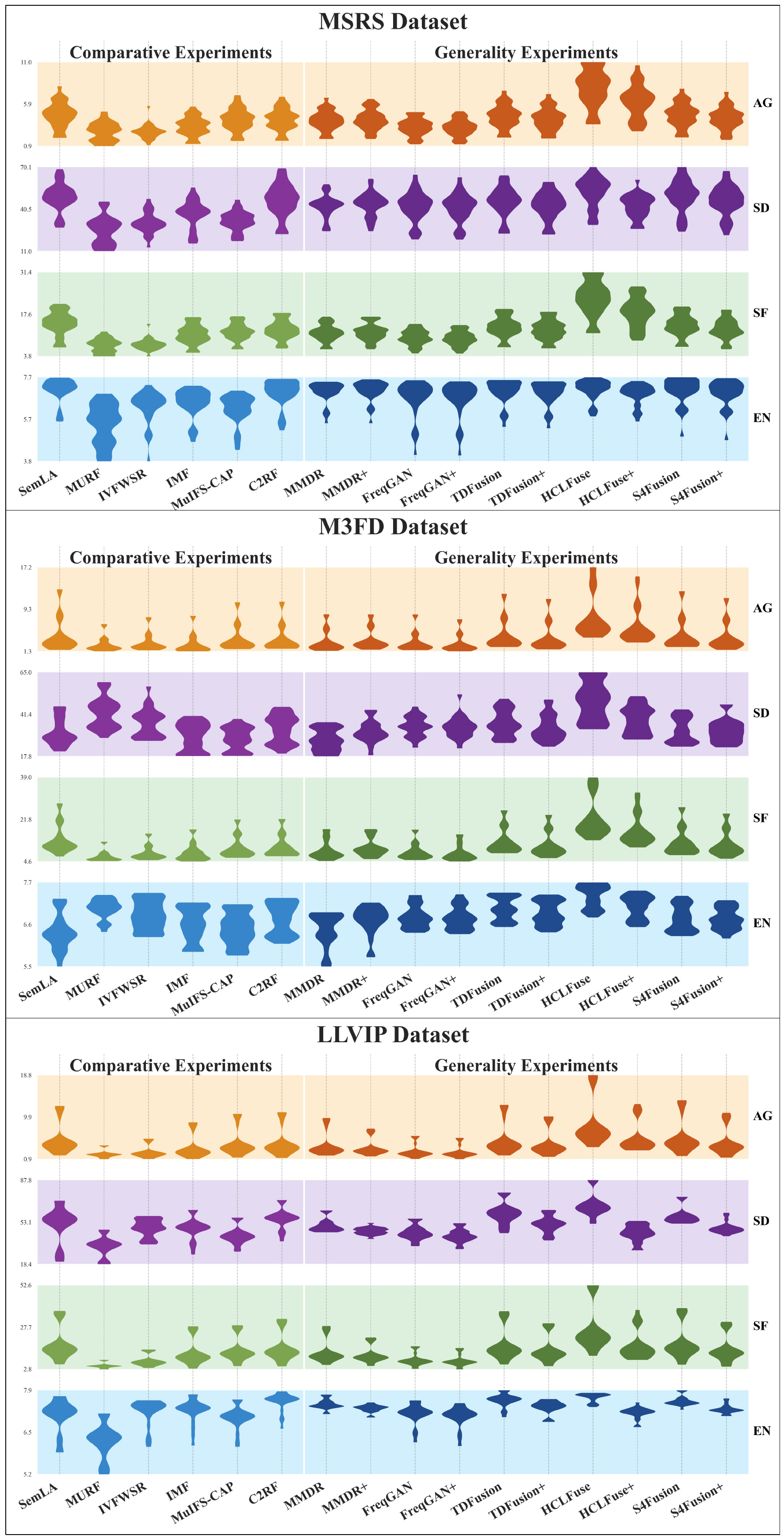}
\caption{ Image quality evaluation of FusionRegister under two scenarios. The method with ``+" denotes the version aligned via FusionRegister.
}
\label{fig:metric}
\vspace{-0.6cm}
\end{figure}

%% file: exp.tex
\section{Experiments}
\label{sec:exp}

\begin{figure*}[!tbp]
\centering
\includegraphics[width=1\linewidth]{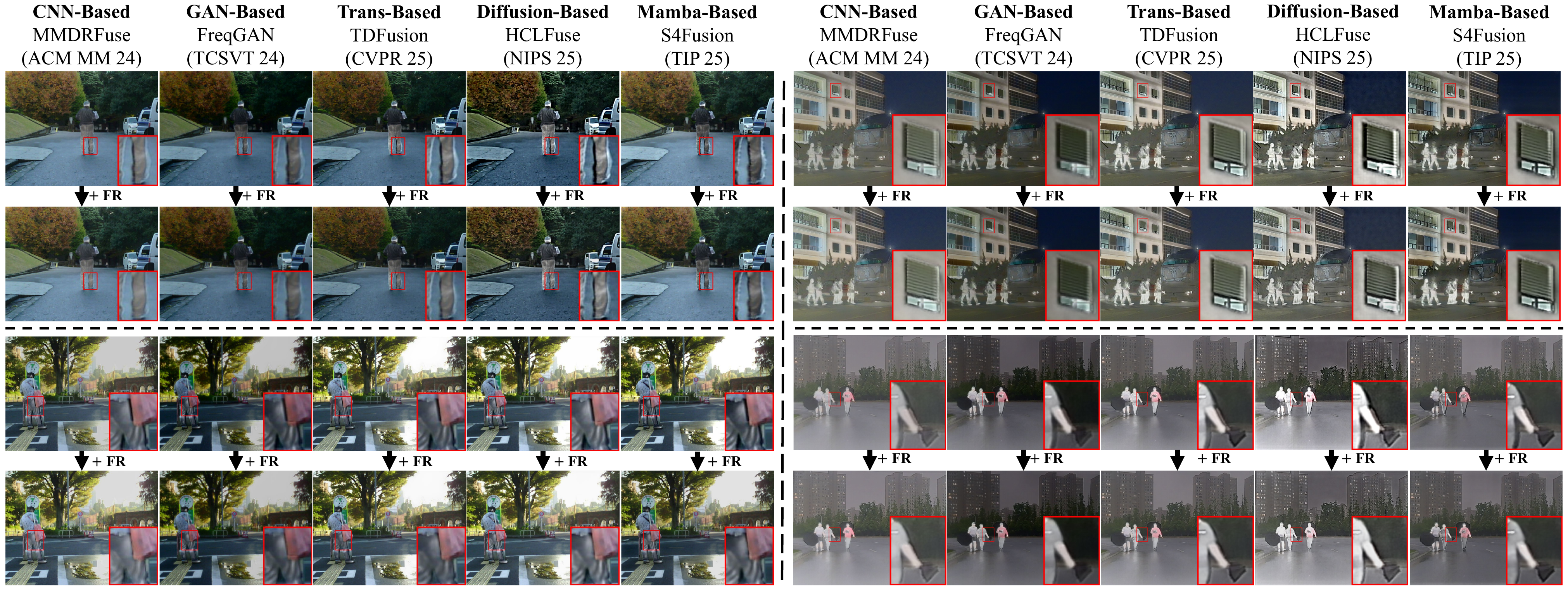}
\caption{As a general framework, FusionRegister successfully completes post-registration for different fusion methods and scenarios, and it completely retains the high-quality fusion performance of the original methods. More qualitative experiments are provided in the supplementary material.
}
\vspace{-0.4cm}
\label{fig:genera}
\end{figure*}

\subsection{Implementation Details}
\label{sec:implementation details}
The proposed method, FusionRegister, is trained on a new multi-modal registration dataset which is collected from the public fusion dataset (MSRS~\cite{seafusion} and M$^3$FD~\cite{tardal}). From these two fusion datasets, we manually cropped 1,333 fully registered $260 \times 260$ patches (426 from MSRS, 907 from M$^3$FD). For testing, 27 MSRS, 21 M$^3$FD, and 20 LLVIP~\cite{jia2021llvip} samples with natural misregistration are used, evaluating at full resolution without synthetic deformation.

To simulate real-world misregistration, we apply random affine transformations to $I_{ir}$: rotation $\left[-2^{\circ}, 2^{\circ}\right]$, translation $\left[-2, 2\right]$ pixels, scaling $\left[0.95, 1.08\right]$, and optional geometric operations (flips or $90^{\circ}$ rotations) with $50\%$ probability.

In the training phase, Adam ($\beta_1=0.9$, $\beta_2=0.999$) with Cosine Annealing (lr: $2\times10^{-4}$ to $1\times10^{-6}$) is utilized. The batch size, patch size and epoch are set to 20, $256\times256$, and 5000. Hyperparameters in loss function are set as follows: $p=1$, $s\in\{1,3\}$,
$\lambda_1=10$, $\lambda_2=1$, $\lambda_3=0.1$, $\lambda_4=10$. All experiments are implemented in PyTorch on RTX 4090 GPU.


\subsection{Evaluation Metrics}
The testing environment simulates real-world conditions, where the imaging device maintains a fixed viewpoint and only coarse cross-modal registration is available, leaving residual misalignment and no ground-truth reference $I_{gt}$. Evaluation is performed from two perspectives: fusion quality and registration accuracy.

For fusion quality, four no-reference metrics are adopted: EN~\cite{EN} (information richness), SF~\cite{sf} (texture sharpness), AG~\cite{ag} (edge strength), and SD~\cite{sd} (contrast). Higher values across these indicators reflect superior visual quality of the fused image.

Registration accuracy is assessed using the Segment Anything Model (SAM)~\cite{sam}, which performs panoramic segmentation on fused results to extract object masks. The overlap and consistency between modalities are then quantified by Intersection over Union (IoU) and PR (harmonic mean of precision and recall), where higher scores indicate better alignment. Since M$^3$FD and LLVIP do not have segmentation labels and MSRS only has coarse annotations, manual labeling is applied to ensure fair evaluation across datasets.

\begin{figure*}[!tbp]
\centering
\includegraphics[width=1\linewidth]{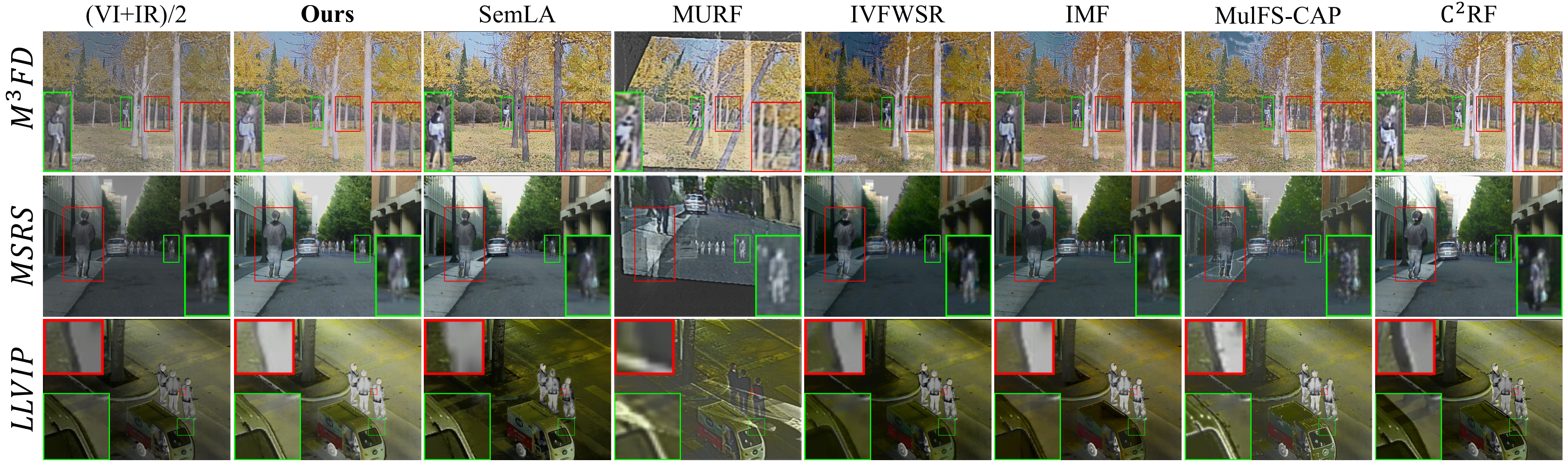}
\caption{ A comparative experiment between FusionRegister and existing registration-based fusion methods.  More visual comparisons are provided in the supplementary material.
}
\vspace{-0.2cm}
\label{fig:comparsion}
\end{figure*}

\subsection{Generality Experiments}
\label{sec:general}
FusionRegister enhances registration accuracy while preserving fusion quality. To evaluate generality, we integrate FusionRegister with five representative fusion models: CNN-based (MMDRFuse~\cite{mmdrfuse}), GAN-based (FreqGAN~\cite{freqgan}), Transformer-based (TDFusion~\cite{gifnet}), Diffusion-based (HCLFuse~\cite{hclfuse}), and Mamba-based (S4Fusion~\cite{s4fusion}), all retrained under identical settings.

As shown in Fig.\ref{fig:genera}, FusionRegister consistently 
maintains the fusion strengths of each method while correcting misregistered regions, demonstrating robustness across diverse scenarios including temporal/spatial misalignments, nighttime scenes, and motion blur. Quantitatively, FusionRegister achieves an average ~5\%  IoU improvement for all fusion method (Table~\ref{tab:metric}), confirming its potential for high-precision applications. The visualization of registration accuracy is shown in the supplementary materials.

Moreover, FusionRegister eliminates ghosting artifacts by consolidating misaligned edges into coherent structures, improving structural consistency despite minor pixel-level metric variations. Fig.\ref{fig:metric} shows post-FR results exhibit lower variance and greater stability across all frameworks. Additional visual comparisons are provided in the supplementary material.

\subsection{Comparison Experiments}
We compare six publicly available infrared–visible registration fusion methods: SemLA \cite{semla}, MURF \cite{murf}, IVFWSR \cite{ivfwsr}, IMF \cite{Wang_2024_IMF}, MulFS-CAP \cite{cap}, and C$^2$RF \cite{c2rf}. As shown in Fig.\ref{fig:comparsion}, most existing approaches fail to adapt to deformation-free inputs or unseen datasets. SemLA and IMF, though capable of limited alignment, often collapse under low-texture or nighttime scenes, yielding blurred or unregistered results. MulFS-CAP performs well only on its training dataset (LLVIP) but generalizes poorly to MSRS and M$^3$FD.

In contrast, FusionRegister consistently localizes and refines misregistered regions, achieving stable post-registration even on entirely unseen data. It is the only method that simultaneously preserves both object and scene integrity while ensuring precise alignment. This advantage is validated by SAM-based segmentation in Table~\ref{tab:metric} (visualized in the supplementary material) and quantitative violin-plot analyses (Fig.\ref{fig:metric}), which show that FusionRegister not only improves mean performance across six methods and three datasets but also reduces variance—demonstrating superior consistency, robustness, and adaptability. More visual comparisons are provided in the supplementary material for completeness. 

\begin{table}[ht]
 \caption{Segmentation quantitative results on MSRS, M$^3$FD and LLVIP datasets represent registration accuracy. The \textbf{bold} and \textcolor{red}{red} represent the best and second-best performance. \textbf{+\textbf{FR}} denote the enhanced versions with FusionRegister.}
 \label{tab:metric}
\tabcolsep=0.14cm
\small
\begin{tabular}{
c |
c c | 
c c |
c c 
}
\hline
\cellcolor[HTML]{FFFFFF} & \multicolumn{2}{c|}{\cellcolor[HTML]{FFFFFF}MSRS} & \multicolumn{2}{c|}{\cellcolor[HTML]{FFFFFF}M$^3$FD} & \multicolumn{2}{c}{\cellcolor[HTML]{FFFFFF}LLVIP} \\
{\cellcolor[HTML]{FFFFFF}Method} & IoU & PR & IoU & PR & IoU & PR \\ \hline

\multicolumn{7}{l}{\textit{Registration-based Fusion Methods}} \\
MURF\cite{murf} & 58.3 & 73.3 & 61.3 & 75.1 & 67.3 & 78.4 \\
SemLA\cite{semla} & 78.7 & 88.4 & 75.0 & 85.9 & 81.0 & 89.7 \\
C$^2$RF\cite{c2rf} & 70.5 & 81.5 & 69.4 & 82.6 & 75.4 & 86.1 \\
IVFWSR\cite{ivfwsr} & 64.6 & 79.4 & 70.7 & 83.2 & 75.6 & 86.1 \\
IMF\cite{Wang_2024_IMF} & 78.7 & 88.4 & 77.2 & 87.6 & 81.3 & 89.8 \\
CAP\cite{cap} & 59.1 & 75.6 & 66.7 & 80.6 & 75.5 & 85.9 \\ \hline

\multicolumn{7}{l}{\textit{Methods with the proposed FusionRegister(\textbf{FR})}} \\

\rowcolor[HTML]{F7F7F7} 
MMDR\cite{mmdrfuse} & 83.6 & 91.3 & 76.9 & 87.3 & 83.7 & 90.5 \\
\rowcolor[HTML]{F2F2F2} 

\multicolumn{1}{r}{\textbf{+FR}} & {\color{red}86.5\textcolor{blue}{$\boldsymbol{\uparrow}$}} & \textbf{93.0}\textcolor{blue}{$\boldsymbol{\uparrow}$} & \textbf{81.6\textcolor{blue}{$\boldsymbol{\uparrow}$}} & \textbf{90.2\textcolor{blue}{$\boldsymbol{\uparrow}$}} & {\textbf{86.0}\textcolor{blue}{$\boldsymbol{\uparrow}$}} & \textbf{93.1\textcolor{blue}{$\boldsymbol{\uparrow}$}} \\[2pt] 

\rowcolor[HTML]{F7F7F7} 
FreqGAN\cite{freqgan} & 80.7 & 89.5 & 74.4 & 85.5 & 82.5 & 90.1 \\
\rowcolor[HTML]{F2F2F2} 
\multicolumn{1}{r}{+\textbf{FR}} & 84.1\textcolor{blue}{$\boldsymbol{\uparrow}$} & 91.4\textcolor{blue}{$\boldsymbol{\uparrow}$} & 80.6\textcolor{blue}{$\boldsymbol{\uparrow}$} & 89.4\textcolor{blue}{$\boldsymbol{\uparrow}$} & 83.2\textcolor{blue}{$\boldsymbol{\uparrow}$} & 90.7\textcolor{blue}{$\boldsymbol{\uparrow}$} \\[2pt]

\rowcolor[HTML]{F7F7F7} 
TDFusion\cite{td} & 85.2 & 92.1 & 76.5 & 87.0 & 81.1 & 89.7 \\
\rowcolor[HTML]{F2F2F2} 
\multicolumn{1}{r}{+\textbf{FR}} & \textbf{86.7\textcolor{blue}{$\boldsymbol{\uparrow}$}} & \color{red}92.9\textcolor{blue}{$\boldsymbol{\uparrow}$} & {\color{red}81.0\textcolor{blue}{$\boldsymbol{\uparrow}$}} & {\color{red}89.7\textcolor{blue}{$\boldsymbol{\uparrow}$}} & 82.9\textcolor{blue}{$\boldsymbol{\uparrow}$} & 90.7\textcolor{blue}{$\boldsymbol{\uparrow}$} \\[2pt] 

\rowcolor[HTML]{F7F7F7}
HCLFuse\cite{hclfuse} & 79.2 & 88.7 & 66.2 & 78.6 & 76.9 & 87.3 \\
\rowcolor[HTML]{F2F2F2} 
\multicolumn{1}{r}{+\textbf{FR}} & 83.7\textcolor{blue}{$\boldsymbol{\uparrow}$} & 91.3\textcolor{blue}{$\boldsymbol{\uparrow}$} & 80.4\textcolor{blue}{$\boldsymbol{\uparrow}$} & 89.4\textcolor{blue}{$\boldsymbol{\uparrow}$} & 84.7\textcolor{blue}{$\boldsymbol{\uparrow}$} & {91.6\textcolor{blue}{$\boldsymbol{\uparrow}$}} \\[2pt]

\rowcolor[HTML]{F7F7F7} 
S4Fusion\cite{s4fusion} & 79.9 & 89.2 & 72.5 & 84.5 & 82.8 & 90.6 \\
\rowcolor[HTML]{F2F2F2} 
\multicolumn{1}{r}{\textbf{+FR}} & 85.9\textcolor{blue}{$\boldsymbol{\uparrow}$} & 92.5\textcolor{blue}{$\boldsymbol{\uparrow}$} & 75.7\textcolor{blue}{$\boldsymbol{\uparrow}$} & 86.6\textcolor{blue}{$\boldsymbol{\uparrow}$} & \color{red}85.3\textcolor{blue}{$\boldsymbol{\uparrow}$} & \color{red}92.0\textcolor{blue}{$\boldsymbol{\uparrow}$} \\ 
\hline

\end{tabular}
\end{table}

\subsection{Ablation Experiments}

\subsubsection{Effectiveness of MRB}

\begin{figure}[!tbp]
\centering
\includegraphics[width=1\linewidth]{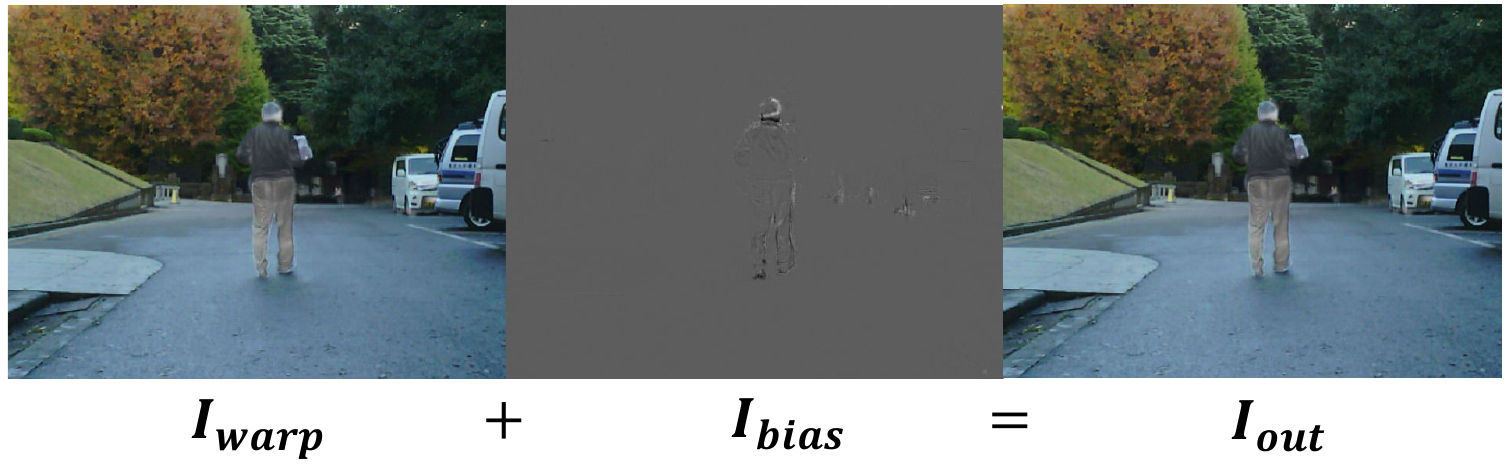}
\caption{ The MRB enables FusionRegister to preserve rich, fine-grained details.
}
\label{fig:bias}
\end{figure}

Transformations based on deformation fields inevitably introduce geometric distortions that degrade structural consistency and fine textures. Without correction, such distortions cause local detail loss in misregistered regions. To validate the necessity of MRB, we performed an ablation study by removing it and re-evaluating fusion results. As shown in Fig.\ref{fig:bias}, the residual map $I_{bias}$ reveals that MRB effectively restores texture details otherwise lost after deformation. The quantitative results in Table~\ref{tab:abl} further show clear declines in multiple image-quality metrics, confirming the crucial role of MRB in detail recovery, structural fidelity, and overall perceptual quality.

\begin{table}[ht]
    \caption{Quantitative analysis of registration accuracy, image quality and complexity from the ablation experiments on MSRS. The \textbf{bold} and \textcolor{red}{red} represent the best and second-best}
    \label{tab:abl}
\tabcolsep=0.13cm
\footnotesize
\begin{tabular}{c|cc|cccc|cc}
\hline
\cellcolor[HTML]{FFFFFF} & \multicolumn{2}{c|}{Reg} & \multicolumn{4}{c|}{Image Quality} & \multicolumn{2}{c}{Efficiency } \\
\cellcolor[HTML]{FFFFFF}Method & IoU & PR & EN & SF & AG & SD & T(s) & P(M) \\ \hline
w/o MRB & 85.6 & 91.9 & 6.92 & 11.46 & 3.95 & 42.71 & \textbf{0.012} & \textbf{2.83} \\
1-d Warping & 85.5 & 92.4 & 7.02 & 11.58 & 4.04 & 43.57 & 0.019 & 2.94 \\
MRB w/ dt & 86.0 & 92.3 & {\color{red}7.03} & 11.53 & 3.96 & \textbf{44.15} & 0.025 & 3.21 \\
MRB w/ dc & 84.8 & 92.1 & 6.98 & 11.41 & 3.98 & 42.52 & {\color{red}0.014} & 3.25 \\
More Layers & {\color{red}86.3} & \textbf{92.9} & \textbf{7.03} & {\color{red}11.65} & {\color{red}4.05} & 43.79 & 0.021 & 7.32 \\ \hline
Ours & \textbf{86.5} & {\color{red}92.9} & 7.03 & \textbf{11.71} & \textbf{4.09} & {\color{red}43.84} & 0.019 & {\color{red}2.94} \\ \hline
\end{tabular}
\end{table}

\begin{figure}[!tbp]
\centering
\includegraphics[width=1\linewidth]{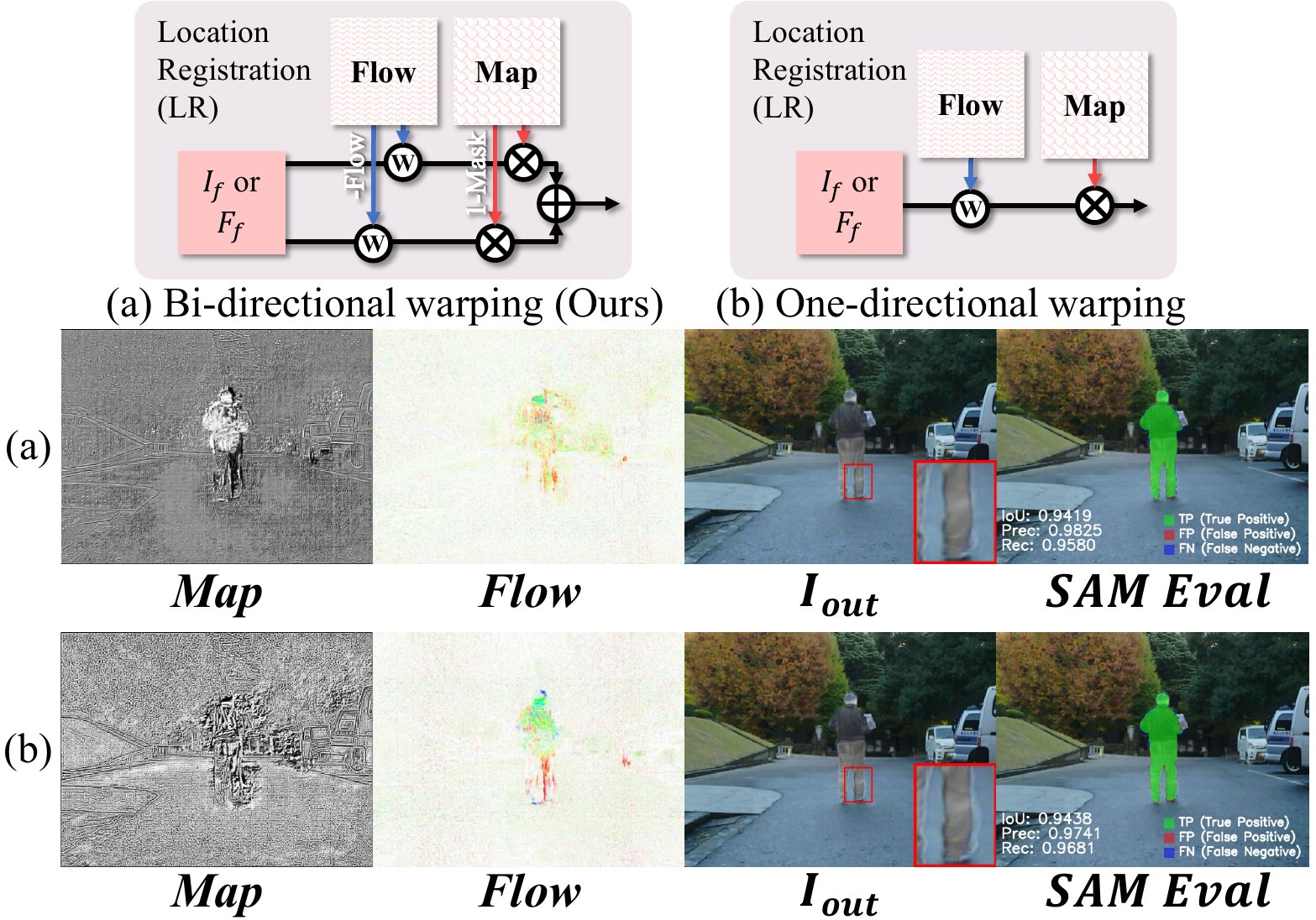}
\caption{Bi-directional warping enables FusionRegister to achieve more precise registration.
}
\label{fig:warping}
\end{figure}

\subsubsection{Effectiveness of Bi-directional warping}


For comparison, we follow other deformation field-based methods that adopt one way warping with a single deformation field. As shown in Fig.\ref{fig:warping} (b), this one-sided strategy overemphasizes suspected misregistration regions and distorts distant structures, leading to noticeable declines in both fusion quality and registration accuracy (Table~\ref{tab:abl}).

In contrast, bi-directional design leverages the mismatch map for explicit region guidance and dual fields for refined geometric correction. As illustrated in Fig.\ref{fig:warping} (a), it preserves structural integrity and recovers local textures more faithfully, confirming the advantage of the proposed mechanism.

\begin{figure}[!htbp]
\centering
\includegraphics[width=1\linewidth]{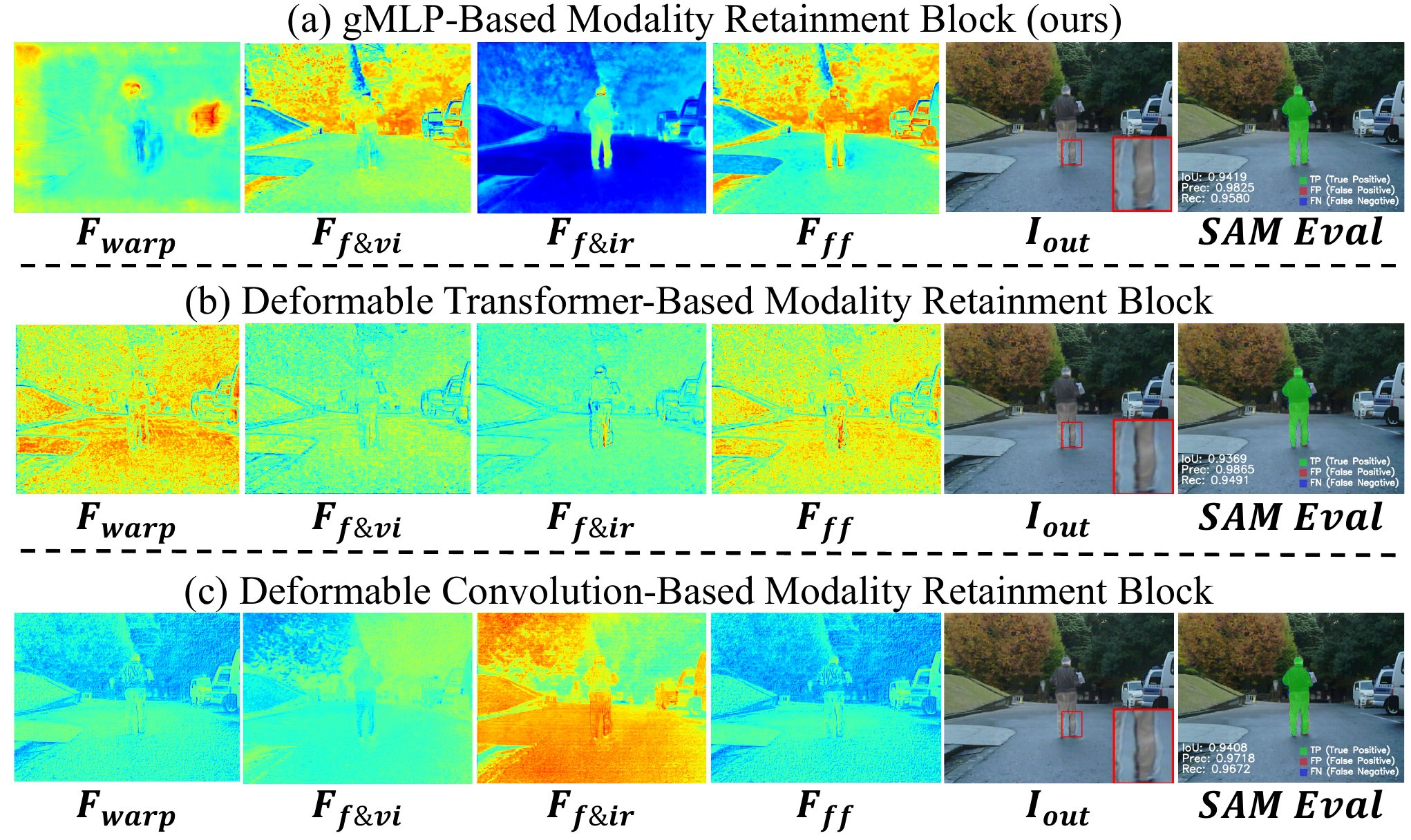}
\caption{ Implementing modality retainment with gMLP leads to the extraction of more comprehensive and abundant features.
}
\vspace{-0.2cm}
\label{fig:gmlp}
\end{figure}

\subsubsection{Effectiveness of Modality Retainment}
The MRB employs a gMLP-based design to preserve modality-specific cues, compared against deformable convolution (DC) \cite{dc} and deformable transformer (DT) \cite{dt}. As shown in Fig.\ref{fig:gmlp}, gMLP-MRB effectively integrates warped features ($F_{warp}$) with modality cues ($F_{f\&vi}$, $F_{f\&ir}$), achieving balanced representation in $F_{ff}$ that maintains both registration accuracy and perceptual fidelity.

In contrast, DT-MRB localizes misregistration but fails to model long-range dependencies, while the limited receptive fields yield of DC-MRB negligible improvement. Although DC offers faster inference, it lacks global modeling capacity. Table~\ref{tab:abl} confirms gMLP-MRB provides the optimal balance of alignment accuracy, detail preservation, and efficiency.

\subsubsection{Effectiveness of Number of Network Layers}
We conduct an ablation study on network depth to balance performance and efficiency. As shown in Table~\ref{tab:abl}, extending from two to three layers yields marginal gains but doubles parameters and increases computation time. Considering the need for lightweight, real-time fusion, we adopt the two-layer architecture. This preserves performance while maintaining compactness, allowing computational resources to focus on misregistration correction and modality retention.

\subsection{Complexity Comparison}
We evaluate methods on two datasets. Table~\ref{tab:time} summarizes computational costs. FR achieves the second-best efficiency. While IVFWSR is fastest, it has the largest parameters and limited adaptability. MulFS-CAP has fewest parameters but slowest inference and poor generalization. FR thus provides superior trade-offs among efficiency, generality, and quality.

\begin{table}[!htbp]
\caption{Comparison of Resource Consumption and Parameters. The \textbf{bold} and \textcolor{red}{red} represent the best and second-best performance. The \textcolor{blue}{blue} represents the \textbf{WORST} performance}
\label{tab:time}
\tabcolsep=0.1cm
\footnotesize
\begin{tabular}{c|ccccccc}
\hline
 & SemLA & MURF & IVFWSR & IMF & CAP & C$^2$RF & \textbf{Ours} \\
\hline
Params (M) & 28.03 & 13.4 & {\color{blue}53.8} & 52.6 & \textbf{1.5} & 10.53 & {\color{red}2.94} \\
\hline
MSRS (S) & 0.75 & 2.23 & \textbf{0.008} & 0.188 & {\color{blue}4.5} & 0.139 & {\color{red}0.019} \\
M$^3$FD (S) & 2.39 & 4.85 & \textbf{0.013} & 0.465 & {\color{blue}10.57} & 0.349 & {\color{red}0.057} \\
\hline
\end{tabular}
\end{table}

%% file: conclusion.tex
\section{Conclusion}
\label{sec:conclusion}
We present FusionRegister, a visual prior guided post-registration paradigm for infrared and visible image fusion. FusionRegister directly refines fused results by localizing and correcting misregistration regions through visual priors. This design enables focused registration, improving efficiency and robustness. 

The current design of FR is limited by the assumption of coarse pre-registration. This limitation prevents FR from achieving satisfactory performance when the inter-modal viewpoint discrepancy is significant. In future work, we will address this critical issue and explore extending the approach to a wider range of multi-modal image fusion tasks.
